\documentclass[journal]{IEEEtran}
\usepackage[utf8]{inputenc}
\usepackage{tabularx}
\usepackage{mathtools}
\usepackage{natbib}
\usepackage{graphicx}
\usepackage{natbib}
\usepackage{amsmath}
\usepackage{array,booktabs,arydshln,xcolor}
\usepackage{lipsum}
\newsavebox\verbbox
\usepackage{url}

\begin{document}

\title{Machine learning models for DOTA 2 outcomes prediction}

\author{Kodirjon Akhmedov
        and Anh Huy Phan
        
\thanks{Kodirjon Akhmedov is with the Department
of Center for Computational and Data-Intensive Science and Engineering (CDISE), Skolkovo Institute of Science and Technology, Moscow,
Russia, 121205 (e-mail: Kodirjon.Akhmedov@skoltech.ru). }% 
\thanks{Anh Huy Phan is with the Department
of Center for Computational and Data-Intensive Science and Engineering (CDISE), Skolkovo Institute of Science and Technology, Moscow,
Russia, 121205 (e-mail: a.phan@skoltech.ru)}

}

\maketitle

\begin{abstract}
Prediction of the real-time multiplayer online battle arena (MOBA) games' match outcome is one of the most important and exciting tasks in Esports analytical research. This research paper predominantly focuses on building predictive machine and deep learning models to identify the outcome of the Dota 2 MOBA game using the new method of multi-forward steps predictions. Three models were investigated and compared: Linear Regression (LR), Neural Networks (NN), and a type of recurrent neural network Long Short-Term Memory (LSTM). In order to achieve the goals, we developed a data collecting python server using Game State Integration (GSI) to track the real-time data of the players. Once the exploratory feature analysis and tuning hyper-parameters were done, our models' experiments took place on different players with dissimilar backgrounds of playing experiences. The achieved accuracy scores depend on the multi-forward prediction parameters, which for the worse case in linear regression 69\% but on average 82\%, while in the deep learning models hit the utmost accuracy of prediction on average 88\% for NN, and 93\% for LSTM models. 
\end{abstract}


\IEEEpeerreviewmaketitle

\section{Introduction}

\IEEEPARstart{T}{he} term Esports is extended as electric sports, which is a form of competition using video games. It is often organized in teams with multiple players, and it can be played by any individual online/offline.  In world competitions, we witness professional players or "pro players" who have gained more experience in particular video games. Esports is not just playing on the computer, but nowadays, it is also considered one of the trending research areas in the scientific community with its predictions and improvement in players' performance [5]. There has been much analysis over the several types of the games such as CS:GO, and Dota 2 the top games under the observation of many Esports researchers [9]. However, there are yet many methods to investigate the games and their outcomes. Finding the best model for predicting the real-time game results is one of the main tasks in Esport [6]. Prediction models are at a high level in many terms. For example, for coaches to improve their players' performance level or for business enterprises, the confident high-level models bring much profit to their budget.

There have been past competitions and investigations over the Dota 2 game prediction. For example, Kaggle’s Dota 2: Win Probability Prediction in collaboration with OpenDataScience is being organized almost every year. We investigated those contests. The most popular articles close to our research area were Win Prediction in Multi-Player Esports: Live Professional Match Prediction [6], To win or not to win? A prediction model to determine the outcome of a Dota 2 match [8], Performance of Machine Learning Algorithms in Predicting Game Outcome from Drafts in Dota 2 [10]. Predicting the winning side of Dota 2 [11] and other articles have been collected and analyzed their results, that still lack some specific proper and easy methodologies which we will discuss them in the related works section.

We can confidently state that scientific research in Dota 2 is still not fully discovered in the science community; therefore, the gap in the knowledge needs more breakthroughs to bring brand new outcomes. While the market is rising at a high rate expected to 1.62 billion USD and 291.6 million people are forecasted to be occasional viewers of Esports by 2024. Thus, the paper has strongly motivated and covered the Dota 2 game analysis. The project covers Esports’ area: develops a real-time data collection process for tracking the in-game data of the gamer using game state integration (GSI), the research focuses upon the Dota 2 Esports Steam game, investigates the multi-forward steps prediction approach using the target that characterizes the success of a player throughout the game process. Real-time prediction of the game or the post-game predictions will be possible with our real-time data collection python tool or GUI and machine and deep learning predictive models to analyze the match result in advance.

The paper has the exact current and long-term goals to achieve for handling the problem statement. We discuss the contemporary state-of-the-art works in Dota 2 match outcome prediction and investigate the possibility of developing machine learning models for the current plans. The following aims of the research are implemented in this paper: \\
\textbf{Contributions}
\begin{itemize}
    \item Investigating how the real-time data can be collected and creating a tool for collecting data from players.
    \item Implementing machine learning and deep learning models to predict the outcome of the match.
    \item Data analysis, investigating features and defining most essential variables, testing models accuracy, and finding considerable features.
    \item Experimenting with players on different occasions and recording results for comparison among several machine learning models.
\end{itemize} 
The following section will give an overall outline of the Dota 2 video game and Steam gaming platform. In section III, we focus on the literature review, where we discuss and explain the limitations of existing papers. In section IV, we propose our new method of collecting real-time data and analyze shortcomings of the data tracking. Section V will be dedicated to theory and algorithms where all explanations of the hypothesis, sensitivity analysis, feature engineering, data prepossessing will be fully explained. Section VI will be dedicated to methodology and experiment setups that define machine learning predictive models and their parameters, compare models, and discuss certain performance levels. Section VII will run through the experimenting where real-time data collection and LR, NN, and LSTM experiments will be presented in tables and graphs and discussed. Section VIII demonstrates how the graphical user interface works and built using PyQt designer and some examples of working GUI. Finally, in the last section, we will conclude what outcomes we got from this paper and tell about long-term future goals and the paper's impact on the innovation.
\section{Dota 2 video game and Steam gaming platform}
\subsection{Dota 2 video game} A Dota 2 (Short for "Defense of the Ancients") is a MOBA video game developed by Valve. The game has ten players in two teams called ‘Dire’ (represented in red color) and ‘Radiant’(represented in green color); there are five players per team, and the idea is to defend your ancient tower and do not let the enemy come towards your side. 
Each match is played on a map split into two sides by a river. Each player chooses a hero from 115 possible heroes (the number of heroes changes depending on Valve’s updates). Heroes are the main characteristics of how the game goes. If the five players have chosen the right combination of the hero, they have a high chance of winning or vice versa; the good gamers specifically take into account the combinations of their hero [8].
The heroes are defined by the strength, agility, and intelligence. There are five couriers (flying horses) which can help five heroes to deliver some items during the game from the your shops and a secret shop (which you can buy many valuable items by visiting with your hero or sending your courier and asking your courier to bring where you want but if courier seen by the enemy during the way to delivery it can be killed and you need to wait to re-spawn your courier and vice versa) as heroes gain more golds during the game. There are not only heroes but creeps that come every half minute in the game; each wave has three Melee creeps and one Ranged creep, and every $10^{th}$ wave Siege creep will also spawn. During the attack, buildings can also throw a fire for the closest enemy to the building; therefore, bearing in mind that not coming close to the tower unless the player does not have extraordinary power. We can see the full illustration of all heroes, creeps, buildings, maps in a screenshot taken from our live game in Figure 1.  \\

\begin{figure}
\centering
\includegraphics[width=0.48\textwidth]{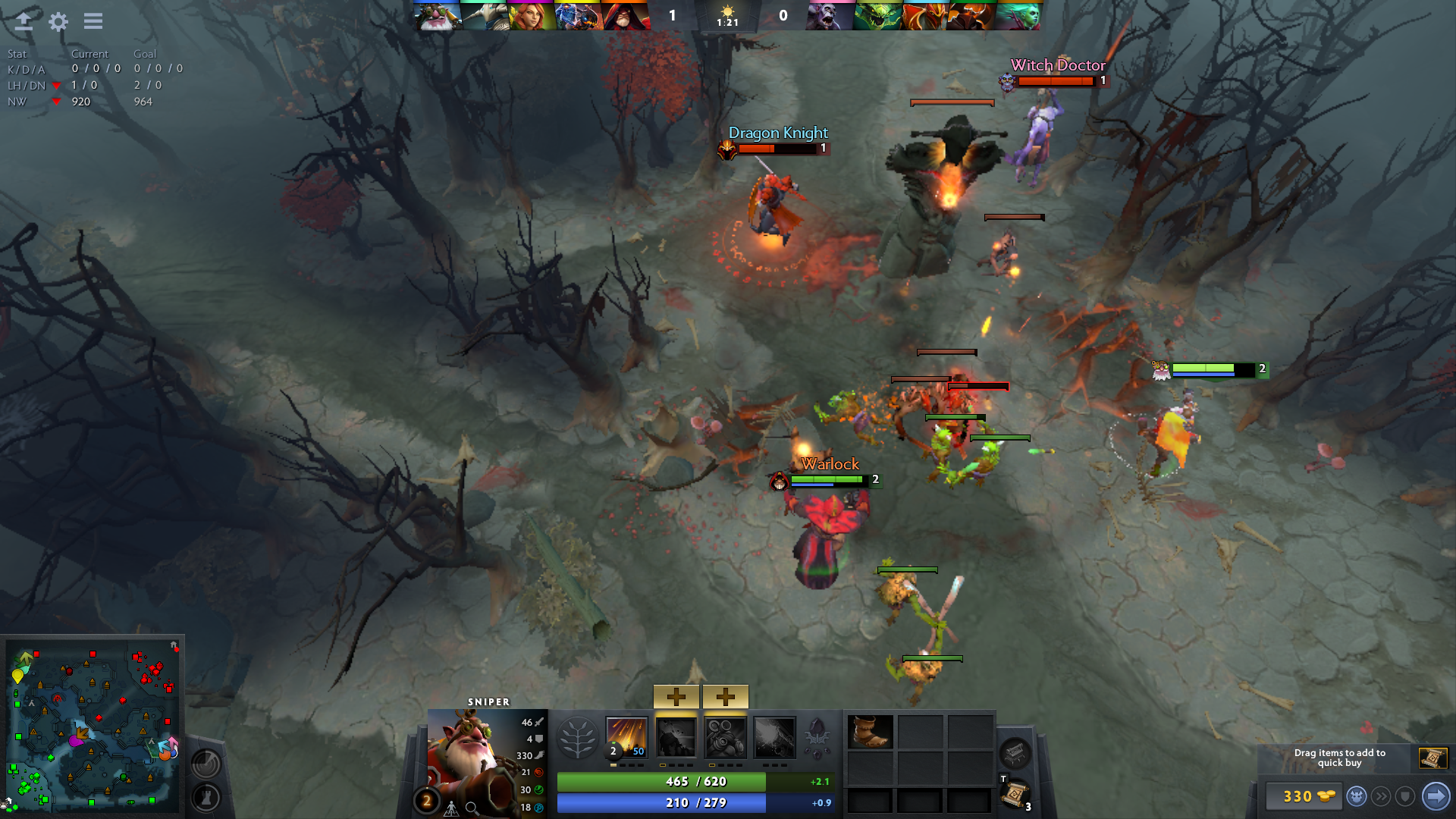}
\caption{Battlefield of the Dota 2. Bottom left corner is the current map, in the center “Radiant” creeps and heroes fighting against “Dire” and tower throwing fire to “Radiant.”
}
\end{figure}

\subsection{Steam gaming platform}
Steam is a video game digital distribution service which is also developed by Valve, where it enables players to connect with each other and play online games at the same time. Steam enables several possibilities for instance video streaming, and social networking, together with updating feature of the games. Each player in the Steam has the ability to have a Steam Token where we use in identifying particular player id. Our research's data collection process is also integrated with Steam Token where we need to get the user’s token to our game state integration (GSI) configuration. 

\section{Related works}
We have researched over the related works, several SOTA articles and different competitions based on Dota 2 win predictions, data tracking, and other related fields have been selected and reviewed for our literature. The authors in [6],  used standard machine learning models, feature engineering, and optimization. Which the method give up to 85\% accurate after 5 minutes of gameplay used 5.7k data from the international 2017 Dota 2 tournament. [12] proposed methods in two steps,  in the first step, Heroes in Dota 2 were quantified from 17 aspects more accurately. In the second step, they proposed a new method to represent a draft of heroes. A priority table of 113 heroes was created based on prior knowledge to support this method. The evaluation indexes of several machine learning methods on this task have been compared and analyzed. 

Another article [8] discussed to improve the prediction results of the existing model, i.e., data collection, feature extraction, and feature encoding. Their idea using augmented regression was defining success probability = (heros picked and success heroes)/(heroes picked) and final success probability = (regression probability + success probability)/2. If the final success probability  $> 0.5$, they predict a "Radiant" win; otherwise, "Dire" win, the metrics they used include how an individual hero contributed and how heroes complemented each other in a match. The results show that the pure logistic regression accuracy, asymptotically approaches 69.42\%, and they conclude that just hero selection plays a vital role in winning or losing the game. In augmented logistic regression, the test accuracy score approached 74.1\%, and authors compared the both models.

Conley and Perry were the first to demonstrate the importance of information from the draft stage of the game with Logistic Regression and k-Nearest Neighbors (kNN) [2]. They got 69.8\% test accuracy on an 18,000 training dataset for Logistic Regression, but it could not capture the synergistic and antagonistic relationships between heroes inside and between teams. To address that issue, the authors used kNN with custom weights for neighbors and distance metrics with 2-fold cross-validation on 20,000 matches to choose d dimension parameters for kNN. For optimal d-dimension = 4 they got 67.43\% accuracy on cross-validation and 70\% accuracy on 50,000 test datasets [2].

The article [10] was also taken into consideration, where authors discussed the comparison of several ML models: Naive Bayes classifier, Logistic Regression chosen to replicate the results of the previous works on their collected datasets and Gradient Boosted Decision Trees, Factorization Machines and Decision Trees were chosen for their ability to work with spare data. AUC, and Log-Loss measured the quality of prediction (Cross-Entropy) on ten-fold cross-validation, the result of AUC was 70\% in XGBoost and 68\% in LogRess and Naive Bayes, the results compared in three normal, high, very high skill sets. Kuangyan Song, Tianyi Zhang, Chao Ma [11] have tried logistic regression to predict the winning side of Dota 2 games based on hero lineups. They collected data using an API provided by the game developer. The results were illustrated using training and testing errors, and a conclusion has been made. 

Paper [7] used the replays in Dota 2 was tested in different classifiers . These articles and our research are in completely different concerning methods. In contrast, we have the ML models, and they have the other methods to analyze their project.  

The interesting article [1] used the ARMA model for investigating time series of change in the health of hero also implemented feature exploration analysis; they achieved slightly greater than 80\% accuracy in predicting the sign of significant changes in health and reasonably accurate results with linear  and logistic regression models.

The paper [4] predicted match results with \textit{xpm, gpm, kpm, lane efficiency, and solo competitive rank}, authors compared SVM and NN. The [3] discussed zone changes, team members distribution, and time series methods. They concluded that MOBA teams behaviour is applicable to team skills. [9] was from the HSE university researchers did consider game statistics and True Skills, which is the similar to Elo rating for chess and they have done for CSGO and Dota 2 with AUC, recall, precision values and compared the results. Wang, W. demonstrated predicting the match outcome depending on the hero draft, with the help of neural networks on GPU, and logistic regression methods with hero drafts [13]. Yifan Yang, Tian Qin, and Yu-Heng Lei tried to predict the matches using logistic regression with replay of data and improved their accuracy from 58.5\% to 71.5\% [14].

From the previous researches, we have found the following shortcomings: 

\begin{itemize}
    \item There is a need for developing deep learning models for better accuracy prediction.
    \item The need for the novel theoretical approach is important to implement most reliable models.
    \item Authors have not thought of developing their own data collection tool.
    \item Authors only limited with source codes while players are not scientists to use codes, thus there is a need for developing user-friendly GUI.
 
\end{itemize} 

Hence our contribution is filling the knowledge gap in the discussed researches and proposing our novel approaches to bring more light to the Esport's Dota 2 win prediction scientific area.

\section{Data Collection and Analysis}

Identifying the player's reaction, movements, real-time interactions could reveal more about the player's results over time. Therefore, it is essential to track such significant information as the game progresses; the D2API python library is useful tool. For example, the following command in python: \begin{itemize}
    \item $import$ $d2api$
    \item $api=d2api.APIWrapper(api$ $key="Your$ $Steam$ $Token")$
\end{itemize}
could give us match details when we have the id of the played match. 
\begin{itemize}
    \item $match$ $details=api.get$ 
    \item $match$ $details(You$ $Match$ $ID)$
\end{itemize}
will load that many details in the variable or the same  $api.get$ $heroes()$ can give the details about the match hero. In practice, it was not possible to store in a file in preferred format. For this particular case, we tried to observe other players' data, and it was not our expectation while we were planning to investigate and view changes of data variable features of our games and extract data in the "CSV" format for further research purposes.  

The idea about writing our codes and integrating between Anaconda and Steam gaming platform was our predominant priority. We discovered the game state integration (GSI) where it could help developers pull data from live Dota 2 games. Python programming language has been selected by our side as a programming language for the development of the data collection tool. We have created a tool that returns us results of real-time data in CSV, JSON, and TEXT files. We have experimented with a quite a lot of our games and successfully collected real-time data using Python as a back-end server. The idea of implementing is vital in researching player’s predictive machine learning and deep learning models or general analysis of players throughout the game. Our practical experiments show that we have developed a tool for collecting data from the Dota 2 MOBA game. The importance of this part of the work is that many researchers want to implement different applications and models for Esports. In contrast, their models could differ, but the tool for collecting data will remain very beneficial for everyone who has a curiosity in in-game and real-time data collection. Thus, this section  can also help researchers and our models by providing a ready and reliable real-time data collection tool. The codes are available at our Github Repository.\footnote{ \url{https://github.com/KodirjonAkhmedov/Real-Time-Data-Collection-Dota-2}}

\subsection{Game State Integration (GSI) in Dota 2}
Game State Integration (GSI) links the game events that are happening with computer, and it is an advantageous method for tracking the real-time data of the playing game. To be more precise, in the case of Dota 2, game state integration has been developed using Python and required configuration files by provided by Valve corporation. Configuration file was the one we need to put inside installed folder of the Dota 2 in our device, for example, in Linux directory has following path address: $/home/test/.steam/debian-installation/steamapps/common/Dota2beta/game/Dota/$ $cfg/gamestateintegration$, where we need to specify our .cfg file inside configurations files of the game.  The GSI consists of an end point section setting which was done for CS:GO but we have created our own endpoint setting for Dota 2 using the Valve's website.\footnote{ \url{https://developer.valvesoftware.com/wiki/Counter-Strike:_Global_Offensive_Game_State_Integration}}:

\begin{itemize}
    \item \textbf{"uri"} which represents the server and port which the game will make post request to this uri: http://127.0.0.1:8080
    \item \textbf{"timeout" "5.0"} The game expects an HTTP 2XX response code from its HTTP POST request, and the game will not attempt submitting the following HTTP POST request while a previous request is still in flight. The game will consider the request to be timed out if a response is not received within so many seconds and re-heartbeat next time with the whole state omitting any delta-computation. If the setting is not specified, then a default short timeout of 1.1 sec will be used.
    \item \textbf{"buffer"  "0.1"} Because multiple game events tend to occur one after another very quickly, it is recommended to specify a non-zero buffer. When buffering is enabled, the game will collect events for so many seconds to report a more significant delta. Setting 0.0 is to disable buffering completely. If the setting is not specified, then default buffer of 0.1 seconds will be used.
    \item  \textbf{"throttle" "0.1"} For high-traffic endpoints, this setting will make the game client not send another request for at least this many seconds after receiving the previous HTTP 2XX response to avoid notifying the service when the game state changes too frequently. If the setting is not specified, then a default throttle of 1.0 sec will be used.
    \item \textbf{"heartbeat" "0.5"} Even if no game state change occurs, this setting instructs the game to request so many seconds after receiving the previous HTTP 2XX response. The service could be configured to consider a game as offline or disconnected if it did not get a notification for a significant period exceeding the heartbeat interval.   
\end{itemize}
\subsection{Real-time data collection and Steam integration}
For Steam platform integration and for collecting data, there is a need to launch Dota 2 game from the Steam gaming platform. Here the server is the python where it works as the “third party” between the game and the computer to enable us to store the data inside the computer directory. The following quick steps can describe the process: 
\begin{itemize}
    \item In the initial step, we define a class called real-time Dota 2 parser. Parser of Dota 2 in real-time takes as input a JSON request and has as output a CSV file. Following that, we define a dictionary for buildings that contains information about buildings, and a function for that variables will be saved all the values and keys pairs given by JSON file to create a dictionary.
    \item In the second step, we took the seven GSI values and their keys, namely, Provider, Map, Player, Hero, Building, Abilities,  and Items.
    \item In the third step, we save a list of values and column names for each arrived request in the dictionary but not buildings because we will save the building values separately and update information about buildings in each post request.
    \item In the fourth step, we save the tracked data to the CSV data. Initially, we create CSV file in our current directory. If it exists, then we renew the data and start writing a new CSV file. An important part to notice there are two cases when our server starts, let us explain with Dota 2 state: the Dota 2 has game states: “pre-game”, “in-progress” and “post-game,” and we define them in our condition when we have these two  “pre-game” and “in-progress,” conditions, then our server starts writing real-time data of the game.
    \item In this fifth step, we define Server Handler by defining several functions of set response and doget and dopost which we will get the size of the data and get the data itself, and we will use the real-time parser that we have defined earlier.
    \item In the sixth step, we begin to write on three different data files: JSON, CSV, and TEXT in each step by making a post request.
    \item In the final seventh stage of our work, we define the HTTP request starting and stopping points where the server works until the user terminates it manually by their kernel inside their environment. In our case was Jupyter Notebook’s stop kernel button. However, this process has been integrated with the Graphical User Interface section, and there is no need to interact with python code. Only start, stop and view buttons enable us to track data easily.
\end{itemize}
\subsection{Data features}
Once we tracked the data we started learning the exciting features and their importance in our machine and deep learning models. The data collection tool enabled us to collect more than real time 160 data features. Lists of game state integration are following, \textit{"buildings", "provider", "map", "player", “hero'',  "abilities", "items", "draft", "wearables"}  where the data has been joined to those main game states including: 
\begin{itemize}
    \item \textbf{“Player”}represents a person who is playing the game and his ability to play the game.
    \item \textbf{“Hero”} represents the player’s choice of hero for the game and hero statistics during the game.
    \item \textbf{“Abilities”} represents the hero’s list of abilities used or how they fluctuated over the time.
    \item \textbf{“Items”} represents the hero’s items and their usage over time.
    \item \textbf{“Buildings”} represents “Dires” or “Radiants” towers (bottom, middle, and top) towers' statistics.
    \item \textbf{“Provider”} represents what is the game which name that is currently in process and what is the game's id Etc. 
    \item \textbf{“Map”} represents the current map, and map states such as does the winner exists or the map is paused Etc. 
\end{itemize}
It is noteworthy that, the columns of our collected data are fixed in all of the games. On the contrary, the rows of the data are in ascending order depending on the game progress. For instance, if the gamers play the game more time, the data capacity will be more, and the row of the data will be greater in length compared to games that are finished in a short period of time.
\subsection{Sensitivity analysis}
Before starting the analysis on the game data features, we need to extract the statistical data from more than 160 features, not all the values are in numerical. More precisely, the collected data could have data types, objects, and integers; therefore, we need to get only the numerical values of data to fit in our model. Thus, we have taken data features whose data values are either \textit{'int16', 'int32', 'int64', 'float16', 'float32', or  'float64'},  extracted statistical data could change depending on how long the game takes place.

Let us define the target variable as y and some independent variables as x. Based on our scientific hypothesis, we have taken the variable that primarily symbolizes the player’s success during the game, that is \textbf{PLAYER.GOLD}, the distribution of the target variable for one game can be seen in Figure 2. \textit{Gold} is the currency used to buy items or instantly revive a hero. Gold can be earned from killing heroes, creeps, couriers, buildings, additional enemies that only stay still in some specific corners of the team’s map side, or by taking bonuses that appear at certain places on the map. It is also passively gained periodically throughout the game. For the input parameter x, we defined ten fixed features with the most considerable correlation to our target variable, Figure 3 illustrates those ten input features' distribution over the game.
\begin{figure}
\centering
\includegraphics[width=0.5\textwidth]{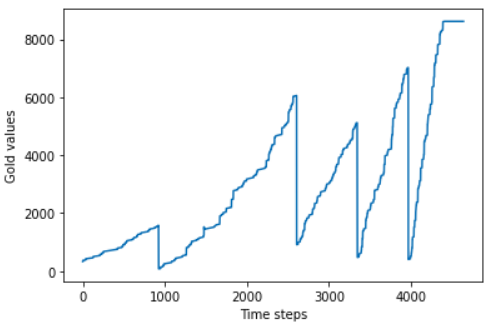}
\caption{ Player gold change for one particular game during the game
}
\end{figure}
\begin{figure}
\centering
\includegraphics[width=0.5\textwidth]{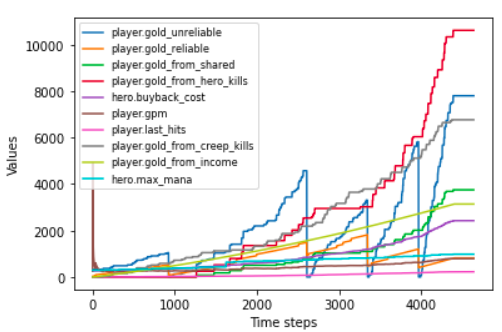}
\caption{Chosen fixed independent features’ change throughout the game in line plot.
}
\end{figure}
\section{Theory and Algorithms}

There has been a hypothesis for predicting the final match result using different techniques and analyses that lacks some decent and easy mechanisms. This section will describe our novel approach, and how the method and algorithms we have tried could be differentiable and important in prediction. As we have seen in our previous section, we have developed a real-time data collection tool using Python as a server for Dota 2 for model implementations, which means our theories and algorithms will be based on our collected data. Depending on our data, we developed our models initially simple linear regression, then the neural network and long short-term memory were trained and calculated the accuracy scores of the models. Moreover, so far, we have done sensitivity analysis on features and their importance. We defined that \textit{player.gold} feature represents the success of the player. With that in mind, we developed a new raw dataset by composing highly correlated $x$ independent features to target $y$ and predicted multi-forward steps. We introduced them as $L$ (time-lagged correlation) and $d$ (multi-forward steps).
\subsection{Time-lagged correlation and multi-forward step prediction}

The method we propose in this paper is dependent on the time-lagged correlation to rows $(L)$ and multi-forward steps $(d)$. Those parameters can manually be adjusted by users, for example, to predict 20 steps ahead of the game while our real-time data is being collected; users can set the $d=20$, then the models will give the results accordingly. Likewise, users can edit the time-lagged correlation $L$ to define the number of composing rows. Let us describe both parameters in detail:
\begin{itemize}
    \item \textbf{Multi-forward step prediction (d)} - represents the number of steps ahead of predicting the game. The parameter plays a crucial role in defining a model's bounds and constructing raw data described in the following subsection during the preprocessing period.
    \item \textbf{Time-lagged (L)} - represents the number of joining rows, imagine the user is tracking data, and user will have initial data which our data tracker is collecting. Imagine each row of that data as variable $x$, which has columns of independent variables $x_1, x_2,...,x_n$ and target $y$ then depending on the $L$, let us set $L=3$, we create a new dataset by composing three rows of the $x$ in one value defined as $X$ in a single row, after that three rows composed, we join next three rows and process continues until $(number$ $of$ $rows(NRow) - (L+d-1))$. In the case of target variable $y$, we add that multi-forward step prediction $d$ to the time-lagged number and write that value to the new dataset and create a raw dataset based on parameters.
\end{itemize}
\subsection{Data construction using model parameters}

We construct new data set using model parameters time-lagged $(L)$ and multi-forward step ahead prediction $(d)$, independent features, target, start indexing, and get our final working data set, consider Table 1 then we have the following data set: 
\begin{table}
\caption{Model construction process using independent variables}
\centering
\begin{tabular}{| c | c | c | c| c | c | c | c|}
\hline
Vector of features  & $x_1$ & $x_2$ & $x_3$ & $x_4$ & $x_5$ & ... & $x_m$  \\
\hline
x(1)      & *  & *  & *  & *  & *  & *   & *   \\
\hline
x(1)      & *  & *  & *  & *  & *  & *   & *   \\
\hline
x(1)      & *  & *  & *  & *  & *  & *   & *   \\
\hline
x(1)      & *  & *   & *  & *  & *  & *   & *   \\
\hline
...       & *  & *  & *  & *  & *  & *   & *   \\
\hline
x(L)      & *  & *  & *  & *  & *  & *   & *   \\
\hline
x(L+1)    & *  & *  & *  & *  & *  & *   & *   \\
\hline
...       & *  & *  & *  & *  & *  & *   & *   \\
\hline
x(NRow-d) & *  & *  & *  & *  & *  & *   & *  \\
\hline
\end{tabular}
\end{table}

$$Dataset=\left(
\begin{array}{lcr}
X(1) & \vert & y(L+d)\\
X(2) & \vert & y(L+1+d)\\
X(3) & \vert & y(L+2+d)\\
X(4) & \vert & y(L+3+d)\\
\ldots & \ldots & \ldots\\
X(NRow-d-L+1) & \vert & y(NRow)
\end{array}
\right)$$
where,

$X(1)=\left(x(1), x(2), \ldots, x(L)\right)$

$X(2)=\left(x(2), x(3), \ldots, x(L+1)\right)$

$X(3)=\left(x(3), x(4), \ldots, x(L+2)\right)$

$X(4)=\left(x(4), x(5), \ldots, x(L+3)\right)$

$\ldots$

$X(k)=\left(x(k), x(k+1), \ldots, x(L+k-1)\right)$

$\ldots$

Eventually, we have $dataset X$ with composed independent variables and $dataset Y$ with the target variable. The shape of our data are entirely depend on the game but for this particular case, $dataset X$ is 3833 rows and 70 columns, and $dataset Y$ consists of 3833 rows and a column with a target.

\section{Methodology and Experimental Setup}
In this section, we will enumerate the languages and programs used in the research, present the metrics and model parameters, describe the implementation of our machine and deep learning models, and compare the different models and several reasons why some models did better compared to others. All models and Graphical User Interface were implemented using Python in Linux Operating System. 
\subsection{Machine Learning models and their parameters}
Three machine learning models have been chosen: \newline
\begin{itemize}
    \item \textbf{Linear Regression} is a supervised machine learning algorithm used to predict values within a continuous range. The idea of the regression problem is dependent on our predicting \textit{target variable}. The dependent variable is player gold and this parameter is not stable during the game, and there are very good linear correlations among certain features therefore, the regression problem has been chosen to predict. We built a linear regression model by splitting our constructed $datasetX$ and $datasetY$ into train and test data, Test size has a 20\% partition, and the rest was dedicated to the training part. We fit the model and take the prediction value from our regressor. The experiments have been done for different datasets, and results were compared in the experiments section. \newline
    \item \textbf{Neural Networks} one of the leading models in time-series forecasting and Esports. Working with artificial neural networks, we could acquire better results due to its ability to train the model through the hidden layers, where each layer takes the input data and processes in the activation functions and passes to the successive layer that can reduce its loss function with forward and backward propagation where biases and weights will be updated and adjusted back and forth during the training process until the number of epochs unless and until the loss value is completely reduced can lead to achieving better model performance.  \newline
    \item \textbf{Long Short-Term Memory} (LSTM) is a type of recurrent neural network (RNN). The output from the previous step feed as an input and the other inputs for the next step and process continues. In LSTM, nodes are recurrent, but all have an internal state. The node uses the internal states as a working memory space that enables information flow over many time steps. Input value, previous output value, and the internal state are all used in the node calculations; the results of the calculations are not only the output value but also used to update the states. Neural networks have parameters that determine how the inputs are used in the calculations, but LSTM has the parameters known as gates that control the node's information. LSTM nodes are rather complicated than regular recurrent nodes, making them learn complex interdependencies and sequences of the input data that results in better outcomes.
\end{itemize}

Several metrics were used to evaluate the performance of the all models. For the accuracy of the linear regression model, we have used the Sklearn library and calculated the $R^2$ but we took final accuracy score as $adjusted$ $R^2$, while NN and LSTM models accuracy scores were taken as $R^2$. \newline
\textbf{Why adjusted $R^2$}, when we have more input variables or predictors are added to a regression model it may surge the $R^2$ value, which entice model's markers to add even more. This is called \textbf{over-fitting} and result in unexpected high $R^2$. Whereas, adjusted $R^2$ is used to identify how trustworthy the correlation and how much is identified by the addition of input features. $R^2$ adjusted accuracy score is evaluated using the sum of squares error (SSE) and the total sum of squares (SST): 
$$R^2 = 1 - \frac{SSE}{SST}$$
where, 
\[ SSE = \sum (y_{test} - y_{predicted})^2 \]
\[ SSE = \sum (y_{testmean} - y_{test})^2 \]
then, 
$$R^2_{adjusted} = 1 - (1-R^2) * \frac{N-1}{N-p-1}$$
where, p is a number of input features and N is a number of rows.

For tuning the optimizers, we chose Adam in NN and LSTM models. We used Adam because it is a replacement optimization algorithm for stochastic gradient descent for training deep learning models. Adam combines the best properties of the AdaGrad and RMSProp algorithms. We used the  Adam parameters with learning $rate=0.01$, $beta_1=0.9$, and $betta_2=0.999$, which were the standard values for the Adam optimization for better performance. The loss function of the LSTM was chosen as the same MSE as the neural network. The model has been trained over 100 epochs since it can update weights and biases of the long training process during 100 epochs, while one epoch means that the optimizer has used every training example once. The batch size was 64, which fits the model in the fastest way possible and returns the output while training with batch size equals to one will take many hours to train our models; therefore, we took the normal value of batch size and evaluated the model performances. The number of total parameters for LSTM training was 622,209, and there were no non-trainable parameters. As for NN, there were a total of 438,913 with 436,225 trainable parameters and  2,688 non-trainable parameters.
\subsection{Architectures of NN and LSTM}
We have tested several architectures for the Neural Network model and calculated the accuracy score for each particular experiment, and we have concluded choosing the architecture that returned us the best performance of the models. We used a sequential model where we can add layers easily. We imported different types of layers to use in our models from the Keras library such that: In our neural networks architecture there are four layers which is performed in Figure 4.

\begin{figure}[!ht]
\centering
\includegraphics[width=0.5\textwidth]{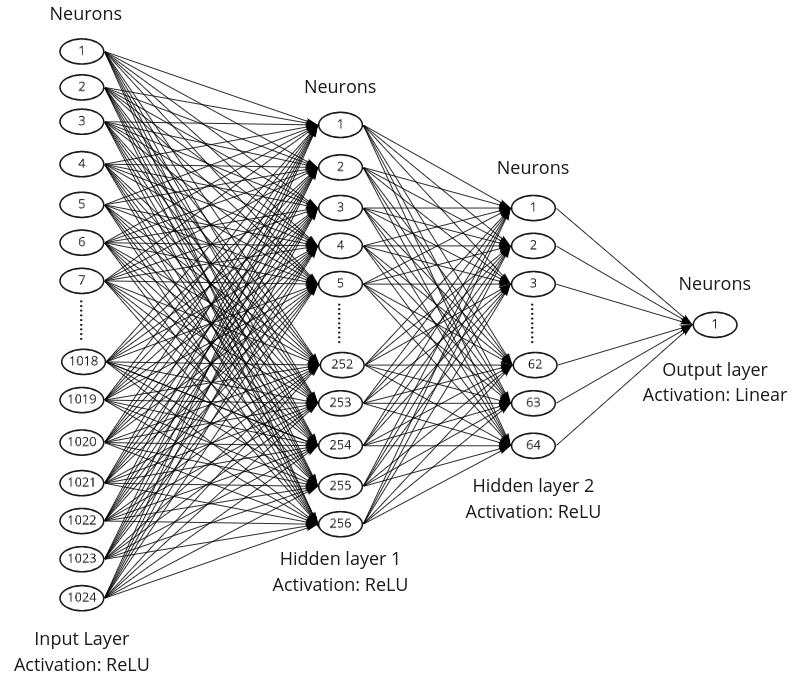}
\caption{Illustration of Neural Network architecture.
}
\end{figure}
In the LSTM model, we used the same sequential model. We standardized the $xtrain$, $ytrain$, $xtest$, $ytest$ with standard and mean and standard deviation as we did in NN. When splitting, we used standard 20\% testing to 80\% training respectively. In comparison with neural networks, we need to reshape the input for enabling the correct working process of the LSTM model. Bearing in mind that, we always have to give a three-dimensional array as an input to the LSTM network, the first dimension represents the samples, the second dimension represents the time-steps, and the third dimension represents the number of features in one input sequence. We imported the LSTM model from TensorFlow Keras layers, and the different architecture of the LSTM has been tested several times, and when the accuracy score performed better results, we chose that architecture as the final working architecture demonstrated in Figure 5.  
\begin{figure} [!ht]
\centering
\includegraphics[width=0.5\textwidth]{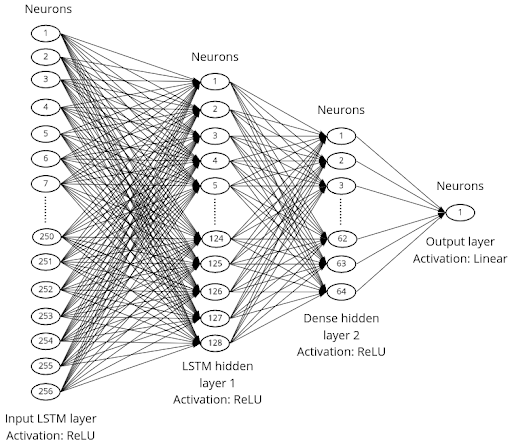}
\caption{Illustration of LSTM architecture.
}
\end{figure}

\section{Results of the Experiments}
This section will describe the result of all tests and experiments that we have done using our models, theory, and algorithms. We would to show the outcomes of the working models and their comparisons in different situations. Overall the experiment process took the final six months of the project period where we were mainly involved in the experiments with different players and collecting their gaming data. 
\subsection{Real-time data collection experiments and results}
Real-time data collection tools were tested in more than 100 games, and all the data has been collected and trained on the models and evaluated the match results. Figure 6 demonstrates the CSV format data that has been collected using our tool. The experiments were done on different values of the $L$ and $d$ where we predicted the low time steps then we increased multi-forward steps respectively, we can see in Table II comparison of three models accuracy performances. Table III describes the increasing $d$ multi-forward steps and fixed $L$ time-lagged correlation experiments on three models.
\newline

\begin{figure}[!ht]
\centering
\includegraphics[width=0.5\textwidth]{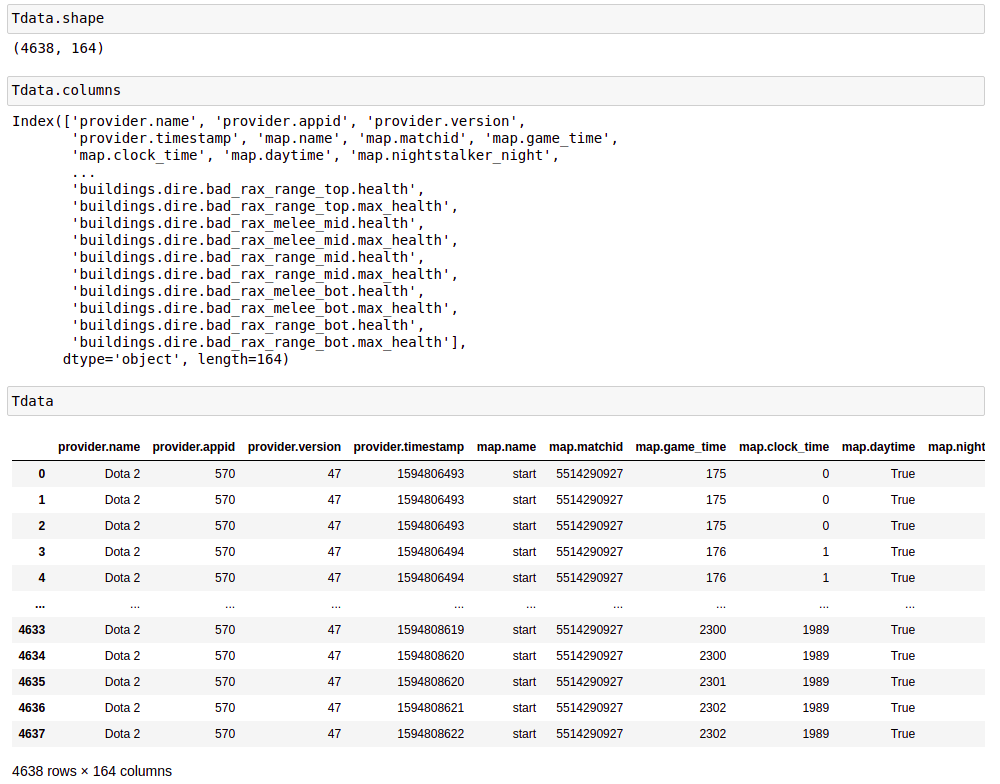}
\caption{Illustration of real-time CSV data collected for a Dota 2 game.
}
\end{figure}

\begin{table}[!ht]
\caption{Accuracy scores comparisons of predictive models}
\centering
\resizebox{\columnwidth}{!}{%
 \begin{tabular}{|c | c| c | c | c | c |} 
 \hline
 Models & d=2, L=5 & d=10, L=7 & d=15, L=10 & d=20,L=15 & d=25, L=20 \\ [0.5ex] 
 \hline\hline
 LR & 0.98 & 0.93 & 0.90 & 0.90  & 0.88 \\ 
 \hline
 NN & 0.99 & 0.94 & 0.93 & 0.93 & 0.91\\
 \hline
 LSTM & 0.99  & 0.95  & 0.94 & 0.95 & 0.94\\  [0.1ex]  
 \hline
 \end{tabular}%
 }
\end{table}

\begin{table}[!ht]
\caption{Accuracy scores comparisons of predictive models with fixed \textbf{L=15} and increasing multi-forward steps}
\centering
\resizebox{\columnwidth}{!}{%
 \begin{tabular}{|c | c| c | c | c | c | c | c | c |} 
 \hline
 Models & d=30 & d=40 & d=50 & d=60 & d=70 & d=80 & d=90 & d=100 \\ [0.5ex] 
 \hline\hline
 LR & 0.86 & 0.92 & 0.74 & 0.67  & 0.64 & 0.68 & 0.67 & 0.70 \\ 
 \hline
 NN & 0.90 & 0.92 & 0.90 & 0.89 & 0.93 & 0.86 & 0.94 & 0.95\\
 \hline
 LSTM & 0.93  & 0.98  & 0.92 & 0.91 & 0.98 & 0.96 & 0.95 & 0.92 \\  [0.1ex]   
 \hline
 \end{tabular}%
 }
\end{table}

\begin{table}[!ht]
\caption{Testing our model on 10 different new game data sets and accuracy scores are following: when d=20, L=15 with 10 fixed input features on all experiments}
\centering
\scalebox{0.76}{
 \begin{tabular}{|c | c| c | c | c | c | c | c| c | c | c|} 
 \hline
 Models & Exp1 & Exp2 & Exp3 & Exp4 & Exp5 & Exp6 & Exp7 & Exp8 & Exp9 & Exp10 \\ [0.5ex] 
 \hline\hline
 LR & 0.81 & 0.69 & 0.75 & 0.72  & 0.84 & 0.82 & 0.96 & 0.87 & 0.85 & 0.87 \\ 
 \hline
 NN & 0.87 & 0.82 & 0.87 & 0.79  & 0.88 & 0.88 & 0.97 & 0.92 & 0.91 & 0.92 \\
 \hline
 LSTM & 0.92 & 0.91 & 0.91 & 0.88  & 0.91 & 0.92 & 0.98 & 0.96 & 0.94 & 0.94\\   
 \hline
 \end{tabular}   
 }
\end{table}

\begin{table}[!ht]
\caption{Testing our model on 10 different new game data sets and accuracy scores are following: when d=20, L=15 with 10 highly correlated input features on all experiments}
\centering
\scalebox{0.76}{
 \begin{tabular}{|c | c| c | c | c | c | c | c| c | c | c|} 
 \hline
 Models & Exp1 & Exp2 & Exp3 & Exp4 & Exp5 & Exp6 & Exp7 & Exp8 & Exp9 & Exp10 \\ [0.5ex] 
 \hline\hline
 LR & 0.80 & 0.75 & 0.78 & 0.70  & 0.67 & 0.80 & 0.96 & 0.82 & 0.87 & 0.88 \\ 
 \hline
 NN & 0.90 & 0.88 & 0.86 & 0.78  & 0.86 & 0.87 & 0.96 & 0.83 & 0.90 & 0.92 \\
 \hline
 LSTM & 0.94 & 0.94 & 0.88 & 0.83  & 0.92 & 0.90 & 0.98 & 0.92 & 0.90 & 0.93\\ [1ex] 
 \hline
 \end{tabular}
 }
\end{table}

\subsection{Graphical visualization of results}
We have performed graphical visualization of our results of predicted and actual values using different types of graphical representations for our consideration, Q-Q plots for viewing the distributions and bar plots and comparison in the separate and merged plots were used to analyse the performances. We can see the plots for different model performances for linear regression in Figure 8. \newline

\textbf{Linear Regression Results} \newline
First, let us consider linear regression results. In Figure 7, we have performed Q-Q plot of the linear regression to see the distribution, and as we can see, the distribution is almost skewed-right. we have achieved accuracy scores depending on the multi-forward prediction parameters, which for the worse case in linear regression 69\% but on average 82\% respectively. \newline

\begin{figure}
\centering
\includegraphics[width=0.48\textwidth]{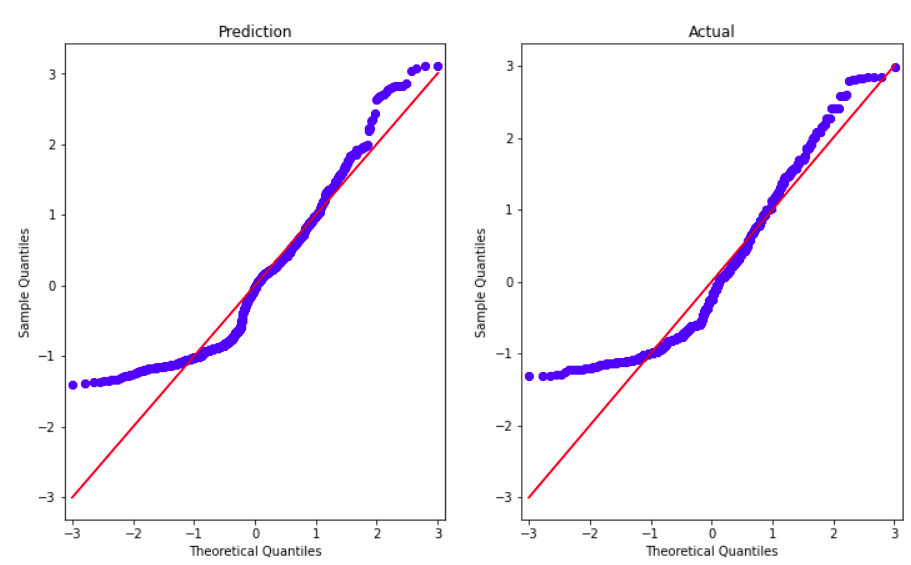}
\caption{LR Q-Q plot of the quantiles}
\end{figure}

\begin{figure}
\centering
\includegraphics[width=0.48\textwidth]{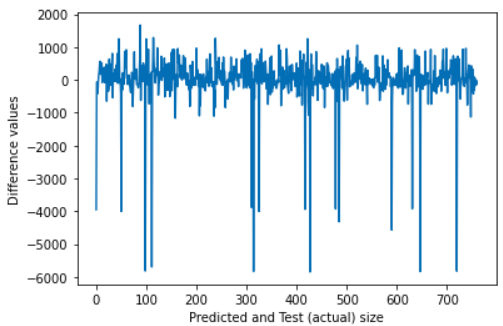}
\caption{Plot of the difference between predicted and actual values of Linear Regression model
}
\end{figure}
\textbf{Neural Network}
In the next experiment, we have tested our neural networks model’s prediction results. The experiment results show better performance for neural network implementation comprising 88\% of average accuracy score for the same data set used in linear regression, which is 6\% higher accuracy score than the LR model. %\newline
\begin{figure}[!ht]
\centering
\includegraphics[width=0.48\textwidth]{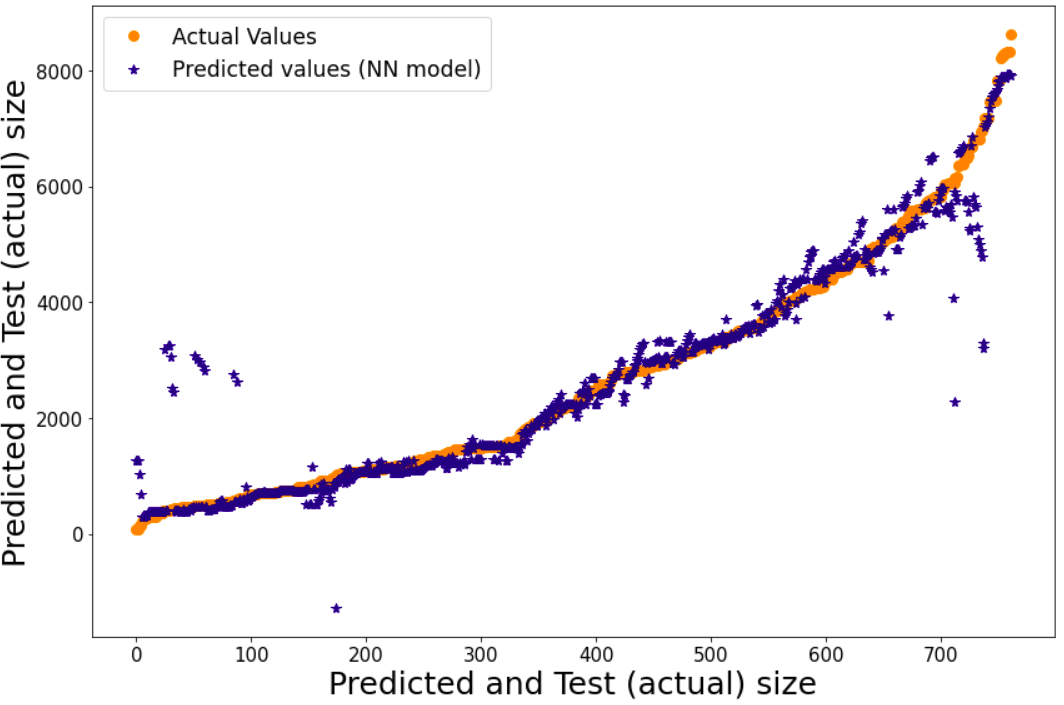}
\caption{Line plot of true sorted (orange) and predicted sorted (blue) values of our target using NN
}
\end{figure}
\newline
We can see some outliers in the data at the beginning and the end of the data set. Considering we are working with large values, we could have such extremum points despite standardization, but there was no considerable impact on the model performance. In the case of neural networks, both predicted and actual values are almost on the same points, representing the high accuracy score with slight non stable values (Figure 9). \newline
\textbf{Long Term Short Memory Results}
The third experiment based on LSTM model, In LSTM model the data for the experiment was the same. The results in this model performed a much better average accuracy score than NN and LR,  93\% accuracy compare to 82\% for LR and 88\% for NN respectively, which means Long Short-Term Memory results are 11\% higher than the Linear Regression model. Reason for such a significant change is that in the LSTM model, biases and weights are updated during the training process over the epochs much sufficiently compared to other two counterparts. There are still some outliers in the prediction such that we have seen some predicted values below the actual values of the target at some points but this model performed perfectly, Figure 10 represents Q-Q plot of quantiles.
\begin{figure}
\centering
\includegraphics[width=0.48\textwidth]{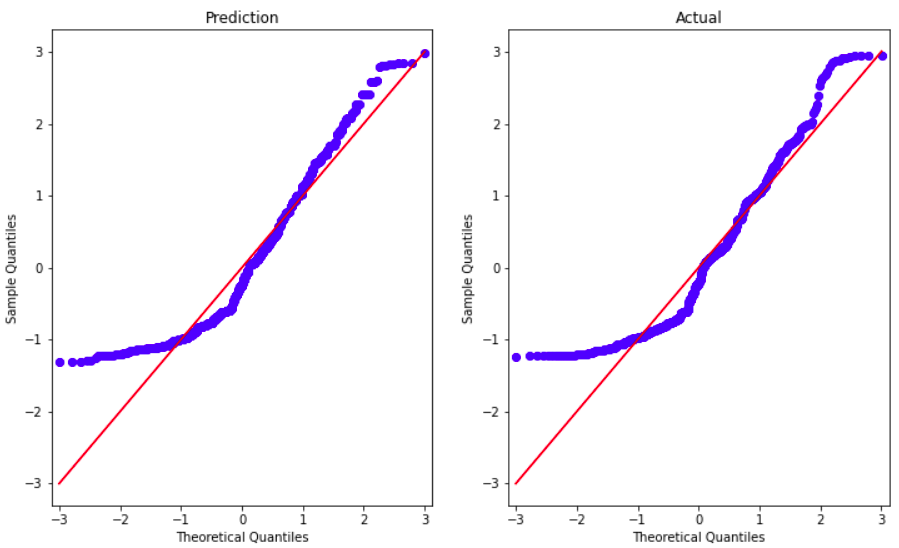}
\caption{LSTM Q-Q plot of quantiles
}
\end{figure} 

Despite experimenting on models accuracy performance, we tested future predictions of our target variable with multi-forward steps $d$ where future values are predicted on three models and compared. We have verified that future prediction values of the target was close to its previously gained values. LSTM and LR were quiet ambitious to have increased prediction while NN showed us slightly decreased prediction based on our experiments (Figure 11).

\begin{figure}[!ht]
\centering
\includegraphics[width=0.48\textwidth]{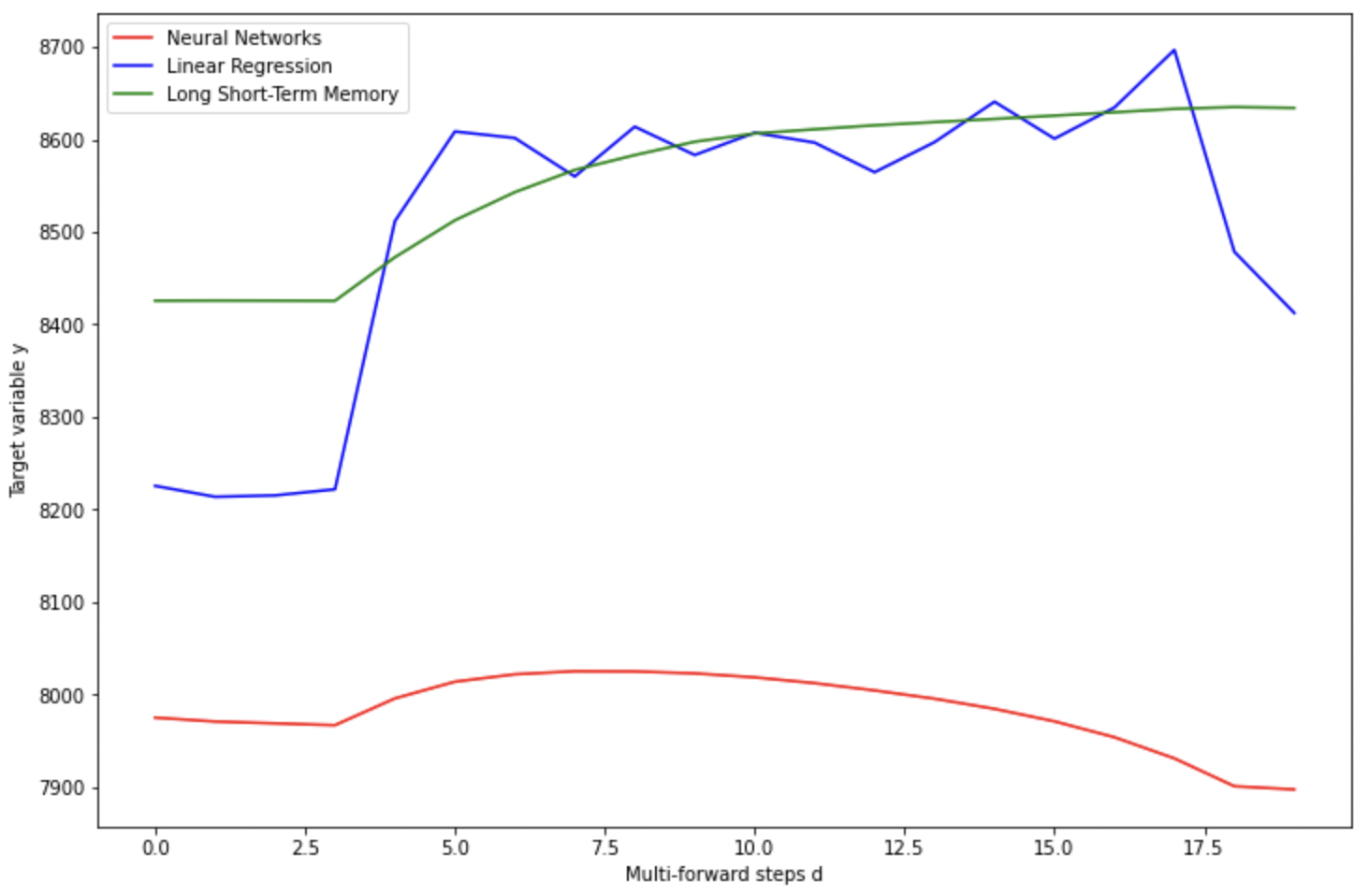}
\caption{Comparison of three models' on $d=20$ multi-forward steps future prediction of target with $L=15$ time-lagged correlation.
}
\end{figure}

\subsection{Discussion of the results among compared models}
Results show that we have achieved considerably decent accuracy scores in all of the models, and the models we tried in some sense gave us a fulfillment which we have attained to see the performance of several models. Initially, we were concerned about the accuracy, but after several experiments, we had a firm belief that the accuracy fluctuates over the range of datasets. However, the range does not drop much from 70\% despite altering several parameters of the model and testing on varying data collected by our data collection tool. 

For example, in deep learning models, LSTM and NN, We deduced that using Neural Networks could give better accuracy than the linear regression model. Moreover, we found that if $d$ and $L$ increase, accuracy decreases but not below 80\% based on our tests, where we have tested with the same dataset and increasing multi-forward steps, and measured accuracy scores, which showed us LSTM also did better performance over Linear Regression and Neural Networks almost in all of the ten experiments except there was better performance in NN when we had chosen fixed composing rows and multi-forward steps, for example, the first experiment they both had the same performances (Table II), but overall, the LSTM was the best compared to the other two models due to its rather  complicated nodes,  making  it learn  complex interdependencies and sequences of the input data that results in better results.

The experiments also included with optimal fixed $L$ model parameter and predicting increasing $d$ multi-forward steps such that we tested from 30 to 100 steps prediction (Table III). We have deduced that Linear Regression can be significantly affected with increase of $d$ where it showed low performance, while NN and LSTM models could handle long steps ahead prediction with accuracy not lower than 85\%.

Our findings of real-time data collection aided us in implementing machine learning models. Data consists of more than 160 columns and more rows depending on game time. The applications of our current discussion are essential in evaluating performance of machine learning models Dota 2 match outcome prediction. In terms of the data collection process, the real-time data collection python script could ease the job of most of the researchers to have ready tool for their models and implement such machine learning models for their different research to enhance the field of the research. There are some limitations, such as research needing further development considering more parameters and bringing some more methods and approaches which we considered in long-term goals with grants, and publications.
\section{Real time GUI using PyQt}
Once we have the working model and data collection tool, how can we make the work more user-friendly was the primary question for us? We wanted the users to click on the buttons and evaluate the accuracy scores of the models based on their choice, and collect the data of their games. The main purpose of this section is to reduce the interaction with source codes and interact with the graphical user interface and evaluate everything using our GUI. For that reason, we thought of creating GUI and connecting our source codes in the backend. There were some more options for doing this task. For example, between web-based and desktop-based GUI, we thought that the desktop-based software is a good option, where we found a handy tool which enables programmers to create user interfaces called PyQt, that is a set of Python bindings for The Qt Company's Qt application framework and runs on all platforms supported by Qt, including Windows, macOS, Linux, iOS and Android. PyQt6 supports Qt v6, PyQt5 supports Qt v5, and PyQt4 supports Qt v4. The bindings are implemented as a set of Python modules and contain over 1,000 classes. \newline

We created the GUI using the PyQt designer application for the Linux operating system. Initially, we had several ideas for the creation of the user interface, but finally, we have made a tool that will be easy to interact with, the instruction of how to use the tool explained in the following subsection.  
\subsection{Instruction of the tool}
In this section, we will go through each step of the usage of our GUI for data collecting and evaluating the models' performances, let us see each step in the following stages:
\begin{itemize}
    \item GUI has three main sections: Data-Collection, Ml models, and Results.
    \item GUI has nine buttons and three text browsers to view model performances.
    \item The first part: data-collection, has “Start,” “Stop,” and “Go to the folder” buttons. Once the game starts, click on it, and it will start collecting real-time data. We can view current data in the “Go to the folder.” This button is made to make sure that data is there and to show whether it is working. Once the game finishes safely close the application and enjoy using collected data.
    \item The second part: Consists of machine learning models that are three models: Linear Regression, Neural Network, Long Term-Short Memory in order to use them, click on the name of the models and wait sometimes while the two deep learning models need time to train over the 100 epochs whereas, Linear Regression will do the quick training.
    \item The third part: results that give us accuracy scores. Once a model has done its part, click one of the (See NN/LSTM/LR Result) buttons, and you can get real-time model performances next to the models in a text box.
    \item To close the GUI, click x safely and finish the work.
\end{itemize}

\subsection{Backend connection with data collection and models’ codes}
For the backend connection of the GUI, we have used the libraries of the \textit{PyQt widgets, QApplication, QMainWindow, QPlainTextEdit, QtCore, QtGui,  and Tkinter} together with the file dialog. Those all the libraries enabled us to implement the backend implementation process of the tool. For each model and data collection section, we linked the main UI, for example, if $pushButton_1.clicked.connect(button$ $name$ $of$ $the$ $GUI)$. Similarly all the other subsections of the push buttons were connected. 

The next stage of the process was creating functions for the .py file reading process and integrating them into our GUI. Overall, nine functions were created such that data collection start a simple $os$ library to read .py file and for stopping the collection $sys.exit()$ was connected. In the case of the models, the same procedure for data collection was implemented with $os$ library. However, each model has one additional function which is responsible for reading the results, which means after the run stage of the model, it returns us TXT file and stores it in a directory then this TXT with results inside was read and performed in the text browser with $ui.textBrowser.setText(file$ $name)$. The process was repeated for the other two models, and this stage has been completed successfully. 
The final working GUI was tested over the game. We can see an example of our games in Figure 12, the data has been collected, and the models' accuracy scores have been evaluated.
\begin{figure}
\centering
\includegraphics[width=0.48\textwidth]{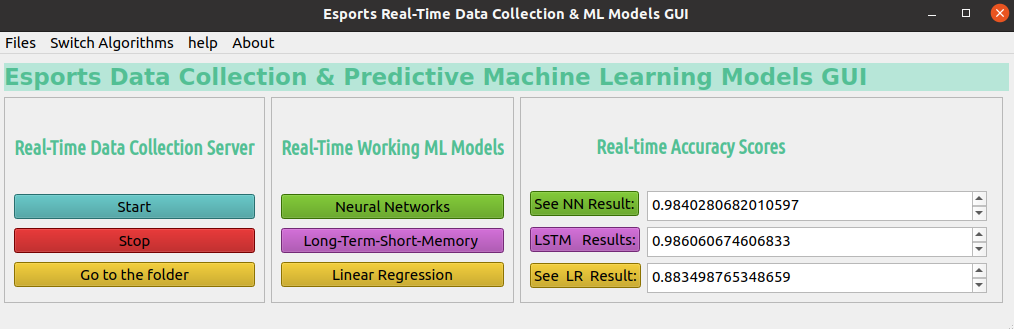}
\caption{Example of working GUI with the accuracy scores of the models on the right bottom
}
\end{figure}
\newline
\section{Conclusion}
In conclusion, all these results suggest that we have built efficient machine learning models from the supervised linear regression model to deep learning models of Neural Networks, and Long Short-Term Memory were implemented to predict game outcomes of Dota 2. It can be stated that using recent approaches, we defined one of the possible ways to achieve our goals. Moreover, with the help of data collection, we successfully implemented our algorithms and investigated essential variables in the Dota 2 game. \newline

Though the experiments run, we have achieved accuracy scores depending on the multi-forward prediction parameters, which for the worse case in linear regression 69\% but on average 82\% while in the deep learning models hit the utmost accuracy of prediction averagely ranging from 88\% for NN to 93\% for LSTM respectively. We have seen that the LSTM model performed better over the other two models in several experiments for flexible occasions of different games. Besides, we have tested dataset on increasing multi-forward steps, and measured accuracy scores, which showed us LSTM also did better performance over Linear Regression and Neural Networks respectively. The performance of the NN was instead relatively better than the linear regression, in which LR was the model with low performance.  \newline

Besides, we had experimented with several steps ahead predictions, as we can deduce that when we have fewer steps ahead prediction and time-lagged correlation, we have high accuracy scores, and vice versa. The graphical user interface section was also designed for the sake of user experience and we illustrated some examples.

\section{References}

[1] ZachCleghern,SoumendraLahiri,Osman Ozaltin,and David L Roberts. Predicting future states in dota 2 using value-split models of time series attribute data. In Proceedings of the 12th International Conference on the Foundations of Digital Games, pages 1–10, 2017.

[2] Kevin Conley and Daniel Perry. How does he saw me? a recommendation engine for picking heroes in dota 2. Np, nd Web, 7, 2013.

[3] Anders Drachen, Matthew Yancey, John Maguire, Derrek Chu, Iris Yuhui Wang, Tobias Mahlmann, Matthias Schubert, and Diego Klabajan. Skill-based differences in spatio-temporal team behaviour in defence of the ancients 2 (dota 2). In 2014 IEEE Games Media Entertainment, pages 1–8. IEEE, 2014.

[4] Petra Grutzik, Joe Higgins, and Long Tran. Predicting outcomes of professional dota 2 matches. Technical report, Technical Report. Stanford University, 2017.

[5] Juho Hamari and Max Sjoblom. What is esports and why do people watch it? Internet research, 2017.

[6] Victoria J Hodge, Sam Michael Devlin, Nicholas John Sephton, Florian Oliver Block, Peter Ivan Cowling, and Anders Drachen. Win prediction in multi-player esports: Live professional match prediction. IEEE Transactions on Games, 2019.

[7] Filip Johansson and Jesper Wikstrom. Result prediction by mining replays in dota 2, 2015.

[8] Kaushik Kalyanaraman. To win or not to win? a prediction model to determine the outcome of a dota2 match. Technical report, Technical Report. Technical report, University of California San Diego, 2014.

[9] Ilya Makarov, Dmitry Savostyanov, Boris Litvyakov, and Dmitry I Ignatov. Predicting winning team and proba- bilistic ratings in “dota 2” and “counter-strike: Global offensive” video games. In International Conference on Analysis of Images, Social Networks and Texts, pages 183–196. Springer, 2017.

[10] Aleksandr Semenov, Peter Romov, Sergey Korolev, Daniil Yashkov, and Kirill Neklyudov. Performance of machine learning algorithms in predicting game outcome from drafts in dota 2. In International Conference on Analysis of Images, Social Networks and Texts, pages 26–37. Springer, 2016.

[11] Kuangyan Song, Tianyi Zhang, and Chao Ma. Predicting the winning side of dota2. Sl: sn, 2015.

[12] Nanzhi Wang, Lin Li, Linlong Xiao, Guocai Yang, and Yue Zhou. Outcome prediction of dota2 using machine learning methods. In Proceedings of 2018 International Conference on Mathematics and Artificial Intelligence, pages 61–67, 2018.

[13] Weiqi Wang. Predicting multiplayer online battle arena (moba) game outcome based on hero draft data. PhD thesis, Dublin, National College of Ireland, 2016.

[14] Yifan Yang, Tian Qin, and Yu-Heng Lei. Real- time esports match result prediction. arXiv preprint arXiv:1701.03162, 2016.

\end{document}